\RequirePackage{fixltx2e}
\documentclass[11pt,english]{article}
\usepackage{mathptmx}
\usepackage[T1]{fontenc}
\usepackage[utf8]{inputenc}
\usepackage[a4paper]{geometry}
\usepackage{fancyhdr}
\usepackage{babel}
\usepackage{array}
\usepackage{amsmath}
\usepackage{amssymb}
\usepackage{setspace}
\usepackage[authoryear]{natbib}
\usepackage[unicode=true,pdfusetitle,
 bookmarks=true,bookmarksnumbered=false,bookmarksopen=false,
 breaklinks=true,pdfborder={0 0 0},pdfborderstyle={},backref=false,colorlinks=false]
 {hyperref}
\hypersetup{
 linkcolor=blue,citecolor=blue,filecolor=blue,urlcolor=blue}

\makeatletter

\providecommand{\tabularnewline}{\\}

\usepackage{pstricks}
\usepackage{epsfig}
\usepackage{multicol} 
\usepackage{caption}
\usepackage{multirow}
\usepackage{tikz}
\usepackage{qtree}
\usepackage{needspace}

\usepackage{ucs}
\usepackage{tikz-qtree}
\usetikzlibrary{trees}
\usetikzlibrary{calc}

\usepackage{etoolbox}
\usepackage{ragged2e}
\raggedcolumns

\tikzset{square arrow above/.style={to path={-- ++(0,.25) -| (\tikztotarget)}}}
\tikzset{square arrow below/.style={to path={-- ++(0,-.25) -| (\tikztotarget)}}}

\let\OLDthebibliography\thebibliography
\renewcommand\thebibliography[1]{
  \OLDthebibliography{#1}
  \setlength{\parskip}{0pt}
  \setlength{\itemsep}{0pt plus 0.3ex}
}

\let\oldmaketitle\maketitle
\renewcommand{\maketitle}{%
  \oldmaketitle
  \thispagestyle{fancy}
}

\makeatother

\begin{document}
\title{Are most sentences unique? An empirical examination of Chomskyan claims}
\author{Hiram Ring\\
NTU Singapore}
\maketitle
\begin{abstract}
A repeated claim in linguistics is that the majority of linguistic
utterances are unique. For example, \citeauthor{Pinker:1994aa} (1994:
10), summarizing an argument by Noam Chomsky, states that ``virtually
every sentence that a person utters or understands is a brand-new
combination of words, appearing for the first time in the history
of the universe.'' With the increased availability of large corpora,
this is a claim that can be empirically investigated. The current
paper addresses the question by using the NLTK Python library to parse
corpora of different genres, providing counts of exact string matches
in each. Results show that while completely unique sentences are the
majority of corpora, actual percentages are highly constrained by
genre, and that duplicate sentences are not an insignificant part
of any individual corpus.
\end{abstract}

\section{Introduction}

The claim that the majority of sentences are unique has been made
in part to highlight the distinctive nature of human language. This
has been a central claim of the ``generative'' (or ``minimalist'')
theory of linguistics, whereby rules are used to generate an ``unlimited
set of sentences out of a finite set of words'' (\citealt{Pinker:1994aa}:
10). Related to this is the claim that ``virtually every sentence
that a person utters or understands is a brand-new combination of
words, appearing for the first time in the history of the universe''
(ibid: 10). This feature of language, along with the ``poverty of
the stimulus'' argument (\citealt{Chomsky:1980aa}), has been used
to suggest that there is some innate language faculty that all humans
are born with (c.f. \citealt{Chomsky:1981aa,Biberauer:2008aa}).

Counterarguments from ``functional'' (or ``usage based'') theories
of language suggest that properties of language arise from general
human cognitive processes applied to communication (c.f. \citealt{Greenberg:1963ul,Pike:1967lr,Hawkins:2014aa}).
The recent success of large language models in mimicking human language
via statistical pattern matching over large datasets of naturalistic
texts seem to support such a usage-based approach. However, if the
Chomsky/Pinker claim regarding the unique nature of utterances is
true, this poses a problem for such usage-based theories.

Large corpora of written or spoken text have become increasingly available
in recent years, allowing for empirical investigation of these claims.
Such corpora are often oriented toward particular domains or genres,
and have been developed to investigate particular linguistic questions,
such as the CHILDES corpora developed to investigate child language
acquisition,\footnote{See https://talkbank.org/childes/} or the British
National Corpus (BNC; \citealt{BNC-Consortium:2007aa}) of 100 million
words, developed ``to represent the full variety of the language''
at the end of the 20th century (with texts collected from 1991-1994).
The large amount of data and the differences in genre allow for systematic
investigation of the degree to which individual sentences are repeated
(or not) in particular contexts. While not fully comprehensive, such
findings can give us insight into the human language faculty.

\section{Methodology}

To investigate this question, several corpora were selected for parsing
via the Natural Language Toolkit (NLTK) library.\footnote{See https://www.nltk.org/nltk\_data/}
This library has been developed in the Python programming language
to allow for rapid investigation of linguistic questions via computational
means (i.e. natural language processing). The library has built-in
readers for multiple corpora, supporting queries such as sentence
retrieval and string matching.

\begin{table}[h]
\noindent \begin{centering}
\begin{tabular*}{1\textwidth}{@{\extracolsep{\fill}}l>{\raggedright}p{5cm}lrr}
{\small{}NLTK corpus} & {\small{}Description} & {\small{}Genre} & {\small{}Texts} & {\small{}Sentences}\tabularnewline
\hline 
 &  &  &  & \tabularnewline
{\small{}brown} & {\small{}A general language corpus (1961) covering a wide range of
styles and varieties of prose. \citep{Francis:1979aa}} & {\small{}balanced} & {\small{}500} & {\small{}57,340}\tabularnewline
{\small{}gutenberg} & {\small{}Selections of public domain books from the Gutenberg text
archive project.} & {\small{}novels} & {\small{}18} & {\small{}98,552}\tabularnewline
{\small{}movie\_reviews} & {\small{}A selection of 1000 positive and 1000 negative movie reviews.
\citep{Pang:2004aa}} & {\small{}reviews} & {\small{}2,000} & {\small{}71,532}\tabularnewline
{\small{}webtext} & {\small{}A collection of diverse, contemporary text genres from web
postings.} & {\small{}mixed} & {\small{}6} & {\small{}25,733}\tabularnewline
{\small{}inaugural} & {\small{}Text of inaugural speeches from US presidents from 1789-2025.} & {\small{}speech} & {\small{}60} & {\small{}5,395}\tabularnewline
{\small{}state\_union} & {\small{}Text of the State of the Union Address by US presidents from
1945.} & {\small{}speech} & {\small{}65} & {\small{}17,930}\tabularnewline
{\small{}bnc} & {\small{}British English from the early 1990s, with imaginative and
informative texts, not restricted to one subject field, register or
genre, both spoken and written. \citep{BNC-Consortium:2007aa}} & {\small{}balanced} & {\small{}4,049} & {\small{}6,026,276}\tabularnewline
{\small{}childes} & {\small{}The English (North American) data from the CHILDES corpus
of transcribed recordings from child language acquisition. \footnotemark } & {\small{}spoken} & {\small{}7,167} & {\small{}2,202,885}\tabularnewline
 &  &  &  & \tabularnewline
\end{tabular*}
\par\end{centering}
\caption{Descriptions of corpora\label{tab:Descriptions-of-corpora}}
\end{table}

\footnotetext{ https://talkbank.org/childes/access/Eng-NA/ }

Each of the corpora was selected to attempt to balance the following
considerations: ease of use (accessibility via NLTK), genre (to sample
as wide a variety as possible), type (to balance spoken vs written),
language (limited to English), and size (at least 5,000 sentences).
Selected corpora are listed in table \ref{tab:Descriptions-of-corpora}
along with brief descriptions.

In this investigation the matching criteria was taken to be ``exact''
string matches. As we are working with written text or written transcriptions
of spoken text, two normalization procedures were implemented. The
first is to lowercase all words, and the second is to remove punctuation.
Since there is no lowercasing or punctuation in spoken communication,
the removal of such conventions allows for a more accurate comparison
between spoken and written corpora. As a final step, sentences with
fewer than three words were excluded. Resulting normalized strings
were then compared.

There are two general measures by which sentences can be compared.
The first is within corpora, and the second is across corpora. Each
of the corpora measured here contained multiple sub-corpora. That
is, the BNC contains multiple documents that represent random samples
of text from various sources, while the CHILDES corpora of North American
English contains documents from various researchers involving transcriptions
of speech from multiple participants. This variation allows for investigation
of patterns that may generalize to a particular domain (i.e. within
child language acquisition) as well as across genres (i.e. patterns
that hold between academic writing and news articles). We can also
observe the effect of corpus size on the prevalence of novel utterances.\footnote{Python code for reproducing these results can be found at: https://github.com/lingdoc/unique\_sentences}

\section{Results}

Table \ref{tab:Counts-of-duplicate} displays the results of an initial
analysis of sentences on the basis of exact string matches. In this
table we can see that for each individual corpus, the average number
of duplicates per text is quite low, with the exception of the CHILDES
corpus, which reaches nearly 8\%. When we compare the total number
of sentences in a corpus with the total number of duplicates (across
texts), we find that this range increases somewhat, from less than
1\% to nearly 30\%, depending on the corpus. If we then compute the
duplicate percentage based on sentences with three or more words for
all sentences across all corpora (nearly 7 million sentences), the
number of duplicates rises to roughly 10\%.\footnote{Thanks to comments and feedback from X/Twitter users on this table
and code to produce it, particularly @tlonic.}

\begin{table}[h]
\begin{centering}
\begin{tabular}{lr>{\raggedleft}p{2.5cm}>{\raggedleft}p{2.5cm}>{\raggedleft}p{2.5cm}>{\raggedleft}p{2.5cm}}
\textbf{Corpus} & \textbf{Texts} & \textbf{Total Sentences} & \textbf{Sentences with 3+ words} & \textbf{Within-text dups (percent)} & \textbf{Across-text/corpus dups (percent)}\tabularnewline
\hline 
 &  &  &  &  & \tabularnewline
brown & 500 & 57,340 & 54,852 & 0.3 & 0.58\tabularnewline
gutenberg & 18 & 98,552 & 90,579 & 1.23 & 1.17\tabularnewline
movie\_reviews & 2,000 & 71,532 & 65,161 & 0.09 & 1.82\tabularnewline
webtext & 6 & 25,733 & 23,360 & 2.96 & 2.98\tabularnewline
inaugural & 60 & 5,395 & 5,353 & 0.07 & 0.19\tabularnewline
state\_union & 65 & 17,930 & 17,370 & 0.19 & 1.06\tabularnewline
bnc & 4,049 & 6,026,276 & 5,429,986 & 1.65 & 5.37\tabularnewline
childes & 7,167 & 2,202,885 & 1,282,614 & 7.92 & 28.42\tabularnewline
 &  &  &  &  & \tabularnewline
\textbf{Totals:} & \textbf{13,865} & \textbf{8,505,643} & \textbf{6,969,275} & \textbf{1.8} & \textbf{9.67}\tabularnewline
 &  &  &  &  & \tabularnewline
\end{tabular}
\par\end{centering}
\caption{Sentences and duplicates in respective corpora\label{tab:Counts-of-duplicate}}
\end{table}

\section{Discussion}

These observations suggest several possible interpretations. One is
that the nature or genre of a text or utterance highly constrains
the likelihood of duplication. Certain genres such as inaugural speeches
are much more highly constrained by the individual (a president) and
the nature of the utterance (a special ceremony installing this individual
as the head of a country), which would likely lead to more unique
utterances. On the other hand, the CHILDES corpus is child-directed
speech whereby caregivers are supporting a child's language acquisition,
leading to a higher likelihood of repetitions as caregivers seek to
scaffold a learner's input (c.f. \citealt{Stoll:2009aa}).

The overall low amount of repeated sentences across corpora (excepting
the CHILDES dataset) does seem to support the initial claim by Chomsky/Pinker,
yet the percentages are not as low as one might suppose from the strength
of the claim (``\emph{virtually every} sentence...''). In fact,
there are duplicate sentences in every single corpus investigated,
indicating that repeated sentences are a natural aspect of language
in use. Given that the CHILDES corpus is the only corpus investigated
which is transcribed solely from spoken interactions, it is possible
that the claim holds for primarily written text, though even with
purely written text corpora it cannot be maintained in its strongest
form.

Work in anthropological linguistics has also shown that many interactions
are ``scripted'' (c.f. \citealt{Frake:1964aa,Hymes:1986aa,Hymes:1989aa}),
such that we would expect to find a higher proportion of duplicated
sentences in everyday life. When ordering a drink at a local shop,
for example, there are likely to be a somewhat limited set of interactional/linguistic
options for participants. Settings or schemas that are more constrained
or often repeated will increase the likelihood of repeated utterances.
It also seems that the degree of duplication may also be affected
by the size of a corpus - higher sentence counts could increase the
likelihood of duplicate sentences. This reflects the nature of ``intertextuality''
(c.f. \citealt{Kristeva:1980aa,Bauman:1990aa,Orr:2003aa,Still:1991aa}),
whereby texts and interactions often refer to each other.

The findings of this study indicate that the strongest version of
the original Chomsky/Pinker claim is false. While the combinatorial
possibilities of language do allow for nearly infinite unique utterances,
the degree to which a given set of sentences consists of unique strings
depends largely on the environment or context in which those sentences
were produced. Repetition is in fact a useful device for framing information
and guiding attention in interaction (consider dialogue or poetry
in scripted performances, for example), and from this perspective
it would be surprising if ``virtually every'' utterance was unique.

There are some caveats that should be made regarding this investigation.
I acknowledge, first, that is somewhat limited in scope. We have looked
specifically at strings of words/tokens that are three or more words
in length, but to what degree can these be considered sentences? There
is some argument to be made that the CHILDES corpus is in fact more
representative of human language given that it is a spoken corpus
of actual interactions, whereas the written corpora are somewhat idealized
representations. Additionally, many ``sentences'' produced in interaction
are in fact phrases or fragments (i.e. ``constructions) - it could
be that breaking sentences into phrases would result in a higher percentage
of duplications.

A second point to note is that this investigation focuses only on
a single language: English. Although it seems likely that similar
principles would hold for other languages, it is possible that these
results would change depending on linguistic and cultural context.
This raises an interesting thought experiment: is there a language
in which investigations of corpora might reveal no duplicate sentences
at all? If this is a possibility, what would such a language look
like and what would it tell us about human language? While such a
question is beyond the scope of this paper, future research could
investigate corpora in other languages to explore these ideas further.

As a final point, the low amount of repeated sentences in the corpora
explored here highlights the creative nature of language. Combined
with the observed increase of repetitions in child-directed speech
corpora, it suggests that there is a functional basis for such repetitions.
This leads to a theoretical observation that repetition (as referred
to by statistical or frequentist models) may be beneficial for language
learning, but does not necessarily accurately reflect how language
is used more broadly. This mirrors the success of recent Large Language
Models, which perform well at recognizable tasks (for which there
is robust training data) but not as well on novel or creative tasks
(for which there is little training data). As such, it seems we might
need a new theory of language that combines elements of the ``generative''
and ``usage-based'' approaches (c.f. \citealt{Rastelli:2025aa}).

\bibliographystyle{linguistics}
\phantomsection\addcontentsline{toc}{section}{\refname}\bibliography{my-bibliog}

\begin{thebibliography}{19}
\providecommand{\natexlab}[1]{#1}
\providecommand{\url}[1]{\texttt{#1}}
\providecommand{\urlprefix}{URL }
\expandafter\ifx\csname urlstyle\endcsname\relax
  \providecommand{\doi}[1]{doi:\discretionary{}{}{}#1}\else
  \providecommand{\doi}{doi:\discretionary{}{}{}\begingroup
  \urlstyle{rm}\Url}\fi
\providecommand{\eprint}[2][]{\url{#2}}

\bibitem[{Bauman \& Briggs(1990)}]{Bauman:1990aa}
Bauman, Richard \& Charles~L. Briggs. 1990.
\newblock Poetics and performance as critical perspectives on language and
  social life.
\newblock \emph{Annual Review of Anthropology}, 19:59--88.

\bibitem[{Biberauer(2008)}]{Biberauer:2008aa}
Biberauer, Theresa (ed.). 2008.
\newblock \emph{The Limits of Syntactic Variation}.
\newblock John Benjamins Publishing.

\bibitem[{{BNC Consortium}(2007)}]{BNC-Consortium:2007aa}
{BNC Consortium}. 2007.
\newblock {British National Corpus 1994, XML Edition}.
\newblock \url{http://hdl.handle.net/20.500.14106/2554}.

\bibitem[{Chomsky(1980)}]{Chomsky:1980aa}
Chomsky, Noam. 1980.
\newblock \emph{Rules and Representations}.
\newblock Oxford: Oxford University Press.

\bibitem[{Chomsky(1981)}]{Chomsky:1981aa}
Chomsky, Noam. 1981.
\newblock \emph{Lectures on government and binding}.
\newblock Dordrecht, The Netherlands: Foris.

\bibitem[{Frake(1964)}]{Frake:1964aa}
Frake, Charles~O. 1964.
\newblock How to ask for a drink in {S}ubanun.
\newblock \emph{American Anthropologist}, 66(6):127--132.

\bibitem[{Francis \& Kucera(1979)}]{Francis:1979aa}
Francis, W.~N. \& H.~Kucera. 1979.
\newblock Brown corpus manual.
\newblock Tech. rep., Department of Linguistics, Brown University, Providence,
  Rhode Island, US.
\newblock \urlprefix\url{http://icame.uib.no/brown/bcm.html}.

\bibitem[{Greenberg(1963)}]{Greenberg:1963ul}
Greenberg, Joseph~H. (ed.). 1963.
\newblock \emph{Universals of Language}.
\newblock Cambridge, Mass.: MIT Press.

\bibitem[{Hawkins(2014)}]{Hawkins:2014aa}
Hawkins, John~A. 2014.
\newblock \emph{Cross-Linguistic Variation and Efficiency}.
\newblock Oxford University Press.
\newblock ISBN 9780199664993.
\newblock
  \urlprefix\url{https://doi-org.remotexs.ntu.edu.sg/10.1093/acprof:oso/9780199664993.001.0001}.

\bibitem[{Hymes(1986)}]{Hymes:1986aa}
Hymes, Dell. 1986.
\newblock Models of the interaction of language and social life.
\newblock In Gumperz, John \& Dell Hymes (eds.), \emph{Directions in
  Sociolinguistics: The Ethnography of Communication}, pp. 35--71. Holt,
  Rinehart and Winston.

\bibitem[{Hymes(1989)}]{Hymes:1989aa}
Hymes, Dell~H. 1989.
\newblock Ways of speaking.
\newblock In Bauman, R. \& J.~Sherzer (eds.), \emph{Explorations in the
  ethnography of speaking}, pp. 433--51. Cambridge: Cambridge University Press.

\bibitem[{Kristeva(1980)}]{Kristeva:1980aa}
Kristeva, Julia. 1980.
\newblock \emph{Desire in Language: A Semiotic Approach to Language and Art}.
\newblock New York: Columbia University Press.
\newblock Trans. Thomas Gora and Alice Jardine and Leon S. Roudiez.

\bibitem[{Orr(2003)}]{Orr:2003aa}
Orr, Mary. 2003.
\newblock \emph{Intertextuality: Debates and Contexts}.
\newblock Cambridge: Polity Press.

\bibitem[{Pang \& Lee(2004)}]{Pang:2004aa}
Pang, Bo \& Lillian Lee. 2004.
\newblock A sentimental education: Sentiment analysis using subjectivity
  summarization based on minimum cuts.
\newblock In \emph{Proceedings of the ACL}.

\bibitem[{Pike(1967)}]{Pike:1967lr}
Pike, Kenneth~L. 1967.
\newblock \emph{Language in relation to a unified theory of the structure of
  human behavior}, Vol. Janua Linguarum, series maior.
\newblock The Hague: Mouton, 2 edn.
\newblock 762 pp. [Second Rev. Ed.].

\bibitem[{Pinker(1994)}]{Pinker:1994aa}
Pinker, Steven. 1994.
\newblock \emph{The Language Instinct: How the Mind Creates Language}.
\newblock New York: William Morrow and Company.

\bibitem[{Rastelli(2025)}]{Rastelli:2025aa}
Rastelli, Stefano. 2025.
\newblock Third-way linguistics: generative and usage-based theories are both
  right.
\newblock \emph{Language Sciences}, 107:101685.
\newblock ISSN 0388-0001.
\newblock
  \urlprefix\url{https://www.sciencedirect.com/science/article/pii/S0388000124000743}.

\bibitem[{Still \& Worton(1991)}]{Still:1991aa}
Still, Judith \& Michael Worton. 1991.
\newblock Introduction.
\newblock In Still, Judith \& Michael Worton (eds.), \emph{Intertextuality:
  Theories and Practices}. Manchester: Manchester University Press.

\bibitem[{Stoll et~al.(2009)Stoll, Abbot-Smith, \& Lieven}]{Stoll:2009aa}
Stoll, Sabine, Kirsten Abbot-Smith, \& Elena Lieven. 2009.
\newblock Lexically restricted utterances in {R}ussian, {G}erman, and {E}nglish
  child-directed speech.
\newblock \emph{Cognitive Science}, 33(1):75--103.

\end{thebibliography}

\end{document}